\documentclass{article}
\usepackage{ijcai26}

\usepackage{times}
\usepackage{soul}
\usepackage{url}
\usepackage[hidelinks]{hyperref}
\usepackage[utf8]{inputenc}
\usepackage[small]{caption}
\usepackage{graphicx}
\usepackage{amsmath}
\usepackage{amsthm}
\usepackage{booktabs}
\usepackage{algorithm}
\usepackage{algorithmic}
\usepackage[switch]{lineno}

\usepackage{bm}

\usepackage{multirow}

\usepackage{hyperref}
\hypersetup{
    colorlinks=true,
    linkcolor=red,
    filecolor=magenta,      
    urlcolor=magenta,
    citecolor=blue,
}

\title{One-Shot Crowd Counting With Density Guidance For Scene Adaptation}

\author{
Jiwei Chen$^1$ \and Qi Wang$^2$ \and Junyu Gao$^2$ \and Jing Zhang$^3$ \and Dingyi Li$^4$ \and Jing-Jia Luo$^1$
    \affiliations Jiangsu Key Laboratory of Intelligent Weather Forecasting and Applications Based on Big Data/ \\State Key Laboratory of Climate System Prediction and Risk Management (CPRM)/\\ ICAR/CIC-FEMD/KLME/ILCEC, Nanjing University of Information Science and Technology$^1$
    \affiliations School of Artificial Intelligence, OPtics and ElectroNics, Northwestern Polytechnical University$^2$
    \affiliations School of Computer Science, Wuhan University$^3$
    \affiliations School of Computer Science and Engineering, Nanjing University of Science and Technology$^4$
}

\begin{document}

\maketitle

\begin{abstract}
     Crowd scenes captured by cameras at different locations vary greatly, and existing crowd models have limited generalization for unseen surveillance scenes. To improve the generalization of the model, we regard different surveillance scenes as different category scenes, and introduce few-shot learning to make the model adapt to the unseen surveillance scene that belongs to the given exemplar category scene. To this end, we propose to leverage local and global density characteristics to guide the model of crowd counting for unseen surveillance scenes. Specifically, to enable the model to adapt to the varying density variations in the target scene, we propose the multiple local density learner to learn multi prototypes which represent different density distributions in the support scene. Subsequently‌, these multiple local density similarity matrixes are encoded. And they are utilized to guide the model in a local way. To further adapt to the global density in the target scene, the global density features are extracted from the support image, then it is used to guide the model in a global way. Experiments on three surveillance datasets shows that proposed method can adapt to the unseen surveillance scene and outperform recent state-of-the-art methods in the few-shot crowd counting. 
\end{abstract}

\section{Introduction}


Crowd scene analysis has attracted the attention of many researchers, driven by its vital applications in smart cities \cite{gao2020cnn} and urban safety \cite{Liu2018Crowd}. Among the fundamental task in this domain, crowd counting in dense environments has emerged as critical research areas within computer vision, garnering substantial scholarly attention due to its difficult technical challenges and significant practical implications \cite{gao2025survey}. 


\begin{figure}[!h]
\centering
   \includegraphics[width=1.0\linewidth]{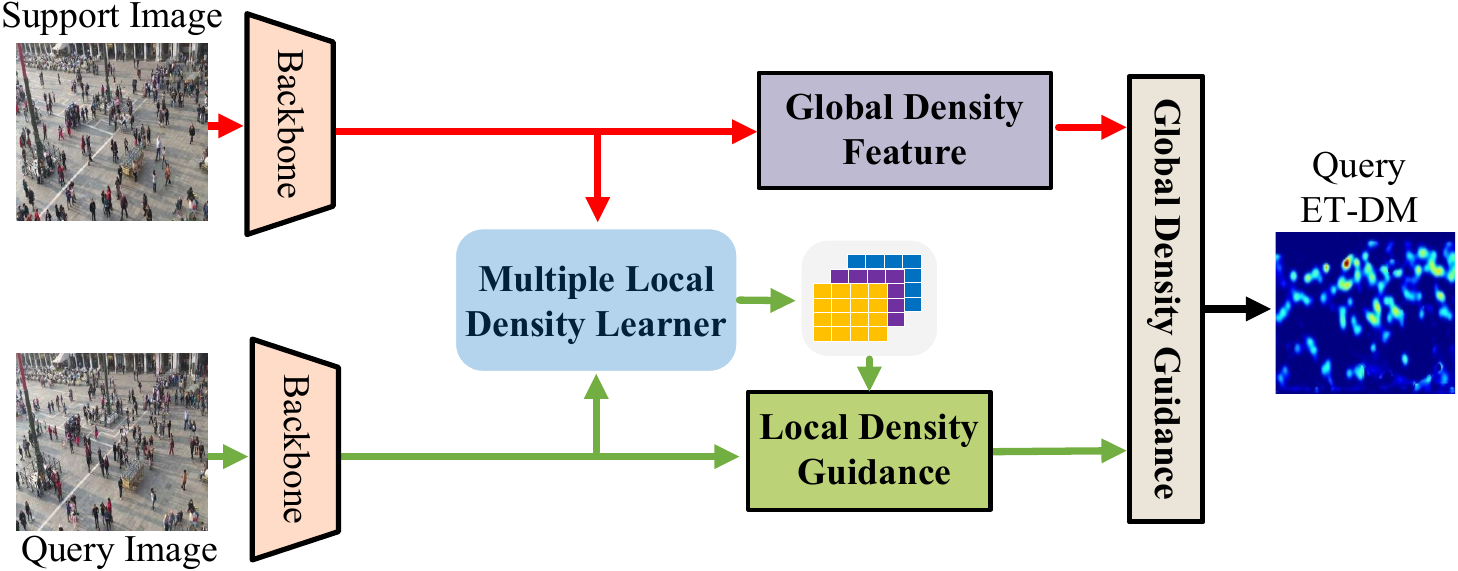}

   \caption{The pipeline of proposed model. Support image and query image are collected from the same category surveillance scene. The multiple local density learner utilizes support and query features to encode these local density similarity matrixes. And they are employed to guide the model in a local way. The global density feature is used to guide the model in a global way. Query ET-DM represents the estimated density map of query image.}
\label{FIG:1}
\end{figure}

Extensive researches have been conducted on crowd counting and achieved great success. These researches can be broadly categorized into two main methods: bounding box-based \cite{li2008estimating,zeng2010robust,benenson2014ten} and density map-based \cite{LiuZPLDL18,babu2018divide,CHEN2019} methods for crowd counting. In the bounding box-based method, the number of positive detections serves as the estimated crowd count. However, in dense crowd scenes, certain pedestrians may fall below the detection threshold due to their small scale. Hence, the density map-based method is proposed, which can record the quantity and location of the crowd, and the crowd count can be derived by integrating the density map. Current state-of-the-art crowd counting researches \cite{CHEN2019,Ranjan_2021_CVPR,chen2025single} predominantly focus on density map-based approaches. Numerous methods have been developed to tackle scale variation \cite{zhang2016single}, complex backgrounds \cite{9161353}, and annotation cost \cite{CHEN2023109506}, etc. Although they have achieved notable success, the problem of scene variation in practical surveillance applications has received relatively little attention. Since surveillance cameras at different locations monitor different crowd scenes, these methods may demonstrate limited effectiveness in the unseen surveillance scene, which represents merely one of thousands of surveillance scenes in smart city applications.

\begin{figure*}[!h]
\centering
   \includegraphics[width=1.0\linewidth]{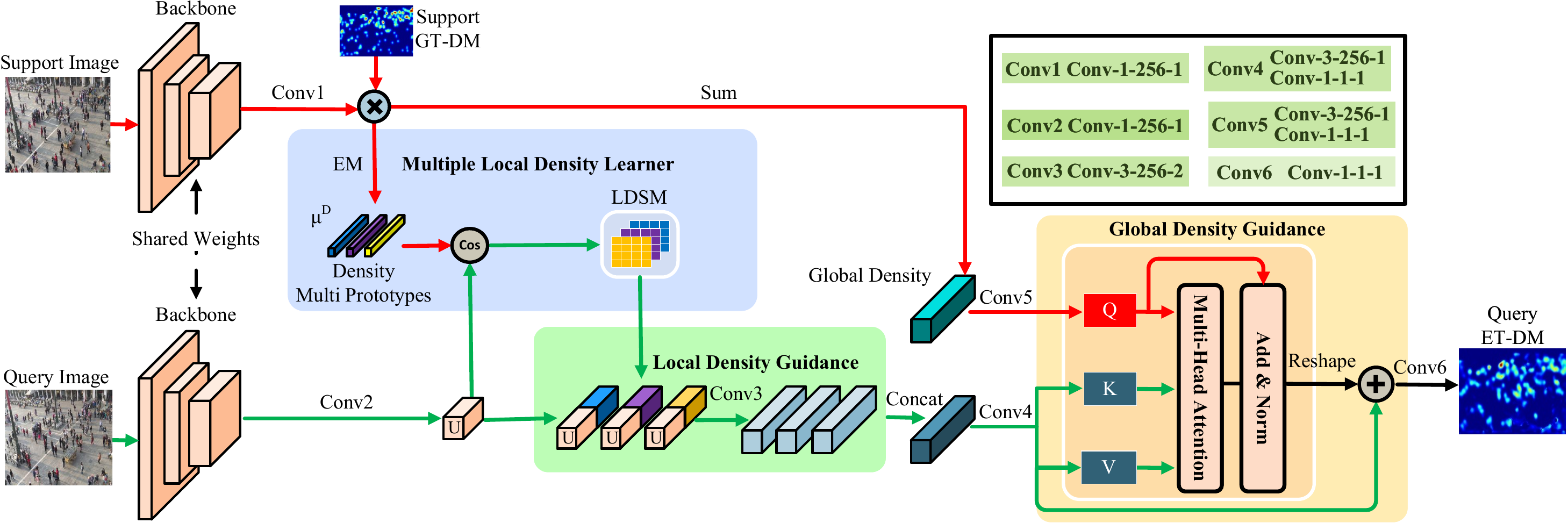}

   \caption{The architecture of the proposed LGD-OSCC (best viewed in color). It employs a dual-branch architecture guided by density features. In the support branch, the ground-truth density map is mapped to support features. On the one hand, the mapping results‌ are employed to encode global density features. On the other hand, they are optimized into density multi prototypes by the EM algorithm in multiple local density learner, which are leveraged to encode these local density similarity matrixes. In the query branch, these local density similarity matrixes are used to guide the model in a local way. Subsequently, transformer utilizes the global density features for global guidance. LDSM represents these local density similarity matrixes. GT-DM represents the ground-truth density map. ET-DM represents the estimated density map. $\rotatebox[origin=c]{45}{$\oplus$}$ represents the element-wise multiplication. \textbf{$\oplus$} represents the element-wise sum.}

\label{FIG:2}
\end{figure*}


To overcome this limitation and make the model adapt to the unseen fixed surveillance scene, we propose treating different surveillance scenes as different category scenes, thereby introducing few-shot learning \cite{song2023comprehensive}. It is well-established that few-shot learning enables models to predict new categories using only a limited number of annotated exemplars. Hence, we propose a \textbf{L}ocally-to-\textbf{G}lobally \textbf{D}ensity-guided \textbf{O}ne-\textbf{S}hot \textbf{C}rowd \textbf{C}ounting (\textbf{LGD-OSCC}) method that enables model adapt to the unseen surveillance scene just using a single labeled image from the target scene. Following metric-learning \cite{zhang2019canet,yang2020prototype,Liu_2022_CVPR} in the few-shot learning, we employ dual density guidance to optimize model performance in the target scene. The framework of our proposed model is shown in Figure \ref{FIG:1}. It integrates a ‌support branch‌, query branch‌, and ‌\textbf{M}ultiple \textbf{L}ocal \textbf{D}ensity \textbf{L}earner (MLDL)‌. The support and query images are collected from the same category surveillance scene, where the support image serves as the one-shot annotated exemplar. Local density features and global density features are respectively extracted from the support feature. Subsequently, the proposed ‌local-to-global density guidance mechanism‌ enables model to adapt to the unseen target scene.

To enable the model to adapt to different density distributions in target image, as shown in the multiple local density learner of Figure \ref{FIG:2}, these support features mapped with the ground-truth density map are adaptively categorized into three multi-prototype representations ‌(high/medium/low-density crowd) by the Expectation-Maximization (EM) algorithm, then the similarity between multi-prototype representation and query feature is encoded into local density similarity matrixes. These ‌density-aware similarity matrixes‌ are respectively‌ fused with query features and activated to provide local density guidance to the model. To further adapt to the global density in the target scene, the global density features‌ are extracted from the support branch to provide global density guidance. In this way, our proposed model achieves precise crowd counting in unseen fixed surveillance scenes. 
  

The LGD-OSCC learning process comprises two phases. ‌In the training phase‌,‌ the base model is firstly trained via loss minimization while freezing the multiple local density learner. Then the base model is fixed, and the multiple local density learner is optimized on the one-shot support image by the EM algorithm. In the test phase‌, the multiple local density learner is firstly optimized on the one-shot support image, then these local and global density characteristics are utilized to guide the fixed base model to adapt to that unseen surveillance scene and predict the crowd counting of query image. 

    In summary, our contributions can be listed as follows.
            
   1. We propose the novel local-to-global density guidance algorithm in one-shot crowd counting to make the model adapt to unseen fixed surveillance scenes.
      
   2. To enhance the generalization, we design the multiple local density learner to extract crowd features with varying densities, subsequently encoding the local density similarity matrix to guide the model in adapting to diverse crowd density distributions of target image in a local way. Concurrently, the global density feature is employed to guide the model in adapting to the global crowd density of target surveillance scene in a global way.
   
   3. Extensive experiments on three benchmark  surveillance datasets validate the generalization‌ of our proposed method, showing its superiority over the state-of-the-art methods in few-shot crowd counting.

\section{Related Work}

\subsection{CLIP-guided Counting}

Recently, large-scale pre-trained models have achieved remarkable success, some researchers propose to leverage CLIP to assist in crowd counting. For instance, to reduce the annotation costs in crowd counting, \cite{liang2023crowdclip} introduce a ranking-based contrastive fine-tuning strategy where crowd patches are mapped into the text prompts. \cite{Paiss_2023_ICCV} drive CLIP by introducing a counting-contrastive loss term, but its effectiveness is significantly diminished for dense objects. \cite{chen2025single} incorporate universal vision-language representations distilled from CLIP into the correlation construction process to overcome domain shifts. \cite{Qian_2025_CVPR} develop the T2ICount framework, which integrates extensive prior knowledge and fine-grained visual comprehension derived from the pretrained diffusion model. \cite{kang2024vlcounter,wang2024language} investigate object enumeration by bridging semantic class labels with visual feature embeddings. Additionally, several open-set methods \cite{ren2024grounding,yao2024detclipv3} demonstrate the strong performance by utilizing CLIP to bridge visual and text representations in interactive object detection. While they have achieved significant success by leveraging CLIP, their proposed method requires substantial computational resources, hindering practical deployment in smart cities.

\subsection{Domain-Generalization Counting}

Domain generalization methods fundamentally rely on the meta-learning framework \cite{li2018learning}, which utilizes meta-training and meta-testing sets to mimic real-world domain shifts during model training and inference, thereby acquiring domain-invariant representations. By eradicating feature correlations, \cite{pan2019switchable,choi2021robustnet} strip away domain-specific stylistic information while preserving unit variance in each feature. \cite{du2023domain} introduce an adaptive sub-domain partitioning method that segments the source domain into multiple sub-domains and iteratively refines them to optimize meta-learning performance. \cite{zhu2023daot} mitigate the inter-domain gap by optimizing the alignment of domain-invariant features between source and target domains using optimal transport. \cite{Peng_2024_CVPR} propose an attention memory bank to extract domain-invariant features, employing a content error mask to remove domain-specific content and an attention consistency loss to ensure memory vector alignment. While their proposed methods significantly mitigate domain shifts across different datasets, they overlook intra-dataset domain shifts arising from varying surveillance scenes.

\subsection{Few-Shot Counting}

Few-shot counting seeks to estimate the number of specific objects using a restricted number of support examples from the target object, e.g., crowd \cite{Ranjan_2021_CVPR}, cars \cite{wen2021detection}, fishes \cite{Ranjan_2021_CVPR}, etc. \cite{Sachin2017,Reddy_2020_WACV,Ranjan_2021_CVPR} investigate the meta-learning mechanism that facilitates parameter adaptation to target class. \cite{2018Few,zhang2019canet,Liu_2022_CVPR} implement a metric-learning method, where a prototypical model embeds images from heterogeneous categories into a metric space, facilitating adaptation to unseen categories and enhancing detection accuracy. To count dense objects, \cite{You_2023_WACV} design the SAFECount which is equipped with the similarity-aware feature enhancement module. To enhance the diversity of training data, \cite{Doubinsky_2024_WACV} modify a pretrained model to utilize caption-based similarities for generating unseen but realistic data, which combines the semantic and geometric features of different training samples. To achieve high recall and precision, \cite{Pelhan_2024_CVPR} combine the strengths of both density-based and detection-based methods.  Despite substantial progress in cross-category counting, limited advancements have been made in the generalization of the same category across diverse scenes. To enable the model to adapt to diverse surveillance scenes within crowds, \cite{chen2024one} propose utilizing multi-prototype representations of density and foreground for guidance. However, local similarity features are not sufficiently exploited, and global features conflate density and foreground cues, resulting in indistinct semantic characteristics that may lead to guidance failure. In this paper, we first extract the similarity features of local density and the salient global density features, then employ a local-to-global guidance technique to adapt the model to unseen fixed surveillance scenes.

\section{Method}



\subsection{Preliminaries}

In the LGD-OSCC setting, we regard crowd scenes captured in different surveillance scenes as different category scenes. In the $M$ different crowd scenes captured by $M$ surveillance cameras at different locations,  $T^{train}_{m1}$ $ (m=1,2, ..., M)$ represents a series of training images collected from the $m_{1}^{th}$ surveillance scene. In the $m_{1}^{th}$ scene, one training image is randomly selected as the one-shot support image, while remaining images $n$ are employed as query images, which can be formulated as $T^{train}_{m1}=\{ T^{query}_{n}, T^{support}_{random}\}$. Throughout the training phase, a support image is stochastically selected at each iteration to drive the model's adaptation to the target scene. In the inference phase, following the few-shot learning paradigm, a single annotated support image remains requisite to facilitate model guidance. Hence, the inference data structure can be formally expressed as $T^{test}_{m}=\{ T^{query}_{n}, T^{support}_{random} \}$. For the annotated crowd data, we employ the normalized Gaussian kernel consistent with \cite{chen2024one} to encode the ground-truth density map. The density is defined as:

\begin{equation}
d(\dot{p})=\sum _{o\in A _{o}}N(\dot{p};\mu _{A},\sigma),
\end{equation} where $\dot{p}$ represents the pixel position. $A_{o}$ represents the complete crowd annotation $o$ of each image.  $N(\dot{p};\mu _{A},\sigma)$ with mean $\mu _{A}$ and isotropic variance $\sigma$ represents the normalized Gaussian kernel, and it is employed to encode the ground-truth density map.


\subsection{Multiple Local Density Learner}


To enable the model to adapt to the target image in local regions, we propose utilizing the local density features of the support image to guide the model. In the collected crowd image data, we observe that certain regions exhibit high crowd density, manifesting as crowded states, while other regions display low crowd density, characterized as sparse states. Hence, the EM algorithm can be employed to group local regions with similar crowd density levels based on density magnitude, thereby extracting these density multi prototypes. 

Based on the above observation and analysis, we design the multiple local density learner. As shown in Figure \ref{FIG:2}, the ground-truth density map is mapped to the support feature $D^{B\times C \times W \times H }$  in the support branch, and the EM algorithm is employed for clustering local features with similar density distributions‌ to encode these density multi prototypes. They are defined as:

\begin{equation}
\begin{aligned}
\begin{split}
p_{\dot{v}}(s_{i}|\theta ) &=\beta_{c}(r)e^{r\mu_{\dot{v}}^{T}s_{i}},  (\dot{v}=1,2, ..., \dot{V}), \\
\beta_{c}(r) &=\frac{r^{c/2 -1}}{(2\pi)^{c/2}I_{c/2-1}(r)},
\end{split}
\end{aligned}
\end{equation}
where $\dot{V}$ denotes the number of prototype vectors, which is set to 3 in this work. After optimizing on the support feature, these three prototypes respectively capture the characteristics of high, middle, and low crowd densities. Each prototype is derived from regions exhibiting similar density distributions in the support image. The normalization coefficient is defined as $\beta_{c}(r)$, with $r$ serving as the concentration parameter. The Bessel function is denoted by $ I(\cdot ) $.

The EM algorithm is utilized to iteratively derive multiple prototypes from the support feature $D^{B\times C \times W \times H }$ through alternating expectation (E-step) and maximization (M-step) phases. In the E-step, the posterior expectation of each sample $s_{i}$ is computed to estimate latent variables.


\begin{equation}
\begin{aligned}
\begin{split}
E_{i\dot{v}} &=\frac{p_{\dot{v}}(s_{i}|\theta )}{\sum_{\dot{v}=1}^{\dot{V}}p_{\dot{v}}(s_{i}\theta )} \\
&=\frac{e^{r\mu_{\dot{v}}^{T}s_{i}}}{\sum_{\dot{v}=1}^{\dot{V}}e^{r\mu_{\dot{v}}^{T}s_{i}}}.
\end{split}
\end{aligned}
\end{equation}

In the M-step, prototype parameters are optimized by maximizing the likelihood of support feature.

\begin{equation}
\mu_{\dot{v}}=\frac{\sum_{i=1}^{I}E_{i\dot{v}}s_{i}}{\sum_{i=1}^{I}E_{i\dot{v}}}, 
\end{equation} where $I=W\times H$, and it denotes the number of samples. $\mu_{\dot{v}}= \{ \mu_{\dot{v}}, \dot{v}=1, 2, ...,\dot{V} \}$ denote estimated mean vectors.

To enable local crowd density features to guide the model in similarity regions, density multi prototype representations which denote different crowd density levels, are subjected to cosine similarity computation with query features, thereby encoding these local density similarity matrixes. And these similarity matrixes can guide the model in a local way. They are defined as follows:

\begin{equation}
\begin{aligned}
\begin{split}
\delta_{\dot{v}} &= cos(p_{\dot{v}},U) \\
&=\frac{p_{\dot{v}}^{T}U}{\left\| p_{\dot{v}}\right\|_{2} \cdot  \left\| U\right\|_{2}},
\end{split}
\end{aligned}
\end{equation} where $p_{\dot{v}} ( \dot{v}\in \{1,2,3\})$ denote these density multi prototypes, and $U$ denotes the query feature.


\subsection{Local-to-Global Density Guidance}


Support image and query image come from the same category‌ surveillance scene. To enable the model to adapt to target scene and achieve precise crowd density estimation for the unseen surveillance scene, we propose a ‌local-to-global density guidance algorithm‌. This algorithm extracts local density features and global density features from the support image, then incorporates ‌local guidance techniques‌ and ‌global guidance techniques‌. Hence, the model can predict accurate density map of the query image collected from the unseen fixed surveillance scene.


\noindent{\bfseries Local Density Guidance.} To exploit local density features for model adaptation in a local way, we select CNN architectures for localized activation and guidance, due to its superior local feature extraction capabilities. As shown in the Local Density Guidance of Figure \ref{FIG:2}, given three encoded local density similarity matrixes ($LDSM_{\dot{v}}, \dot{v}\in \{1,2,3\})$, the query feature is respectively fused with each as:

\begin{equation}
 \hat{U}_{\dot{v}} = Conv(Concat(LDSM_{\dot{v}},U)),
\end{equation} where $U$ denotes the query feature, and $\hat{U}_{\dot{v}}$ denotes these fused feature maps. ‌Activated by the CNN, three local density similarity matrixes are used to guide the model to adapt to the target image within local crowd regions of varying density levels, simultaneously enhancing its robustness to diverse crowd densities. The activated features further contribute to global density guidance.

\noindent{\bfseries Global Density Guidance.} To exploit global density features for model adaptation in a global way, the global density features of the support image are firstly extracted, subsequently, a module proficient in integrating global features needs to be selected. As shown in the support branch of Figure \ref{FIG:2}, the support feature mapped with the ground-truth density map is summed to encode the global density feature. Due to the capacity for global receptive fields, the transformer architecture is selected to activate global density features. The transformer architecture contains three input parameters: Query (Q), Key (K), and Value (V). We define the global density features as Q, and define the local-activation query feature as ‌both‌ K and V. Based on preparations from both aspects mentioned above, the global density guidance can be formally defined as:

\begin{equation}
O = Q+\psi(SM(\frac{QW_{q}(KW_{k})^{T} }{\sqrt{d_{a}}}))\left(VW_{v}\right), 
\end{equation}
where $Q$, $K$ and $V$ are mapped to the global latent space by learnable parameters $W_{q}/W_{k}/W_{v}$. $SM$ denotes the  $softmax(\cdot)$ fucation, and it is used to normalize the attention scores. $\psi(\cdot)$ denotes the linear layer. $d_{a}$ denotes the dimension of the input.

In Equation 7, we can observe that pairwise density similarity is quantified the correspondence scores between the support feature ($Q$) and query feature ($K$) in the global latent space. And these density similarity scores serve as the weights to guide the query ($V$) global latent representations. The underlying principle is that support-query image pairs exhibit higher similarity predominantly in similar density levels; this similarity-driven learning mechanism consequently reinforces the activation response within commensurate levels of crowd density in the query representations. Consequently, the model is more effectively guided by support features through a global conditioning process, culminating in the prediction of query density map.

\begin{algorithm}[tb]
\caption{Learning local-to-global density guidance‌ for one-shot crowd counting.}
\label{alg:algorithm}

\textbf{Input}:\\  \quad The one-shot support image and its ground-truth density map, query image;\\
\textbf{Output}:\\  \quad local density similarity matrixes, the estimated density maps of query image; 
\begin{algorithmic}[1] 

\FOR {(each support image)}          
    \STATE \textbf{Extract} support and query features from the same category surveillance scene;   \\
    \STATE \textbf{Extract} support density features by mapping the ground-truth density map of support image to support features, as shown in Figure 2; \\
    \STATE  \textbf{Learn} the local density similarity matrix;\\
    
    Predict density multi prototypes $\mu^{D}$ on support density features by the EM algorithm defined in Equation 3 and Equation 4;\\

    Encode the local density similarity matrix (LDSM) by calculating the cosine similarity between density multi prototypes $\mu^{D}$ and query features in the Multiple Local Density Learner, as shown in Figure 2;\\

    \STATE  \textbf{Encode} the global density by summing support density features, as shown in Figure 2;\\
    
    \STATE  \textbf{Guide} query features (U); \\ 
    Activate U using the local density similarity matrix (LDSM) in a local way, as shown in Figure 2; \\
    Activate U using the global density features in a global way, as shown in Figure 2; 
    \STATE  \textbf{Predict} the density map of query image.

    \ENDFOR

\end{algorithmic}
\end{algorithm}

\subsection{Architectural Overview and Learning Mechanism}

As illustrated in Figure \ref{FIG:2}, the proposed architecture is an end-to-end network. The training process is formally defined in Algorithm 1. The support density features are encoded by mapping the ground-truth density map to support features. ‌The mapping results‌ are respectively utilized to generate global density feature and local density feature. In the local guidance‌, the density multi prototypes $\mu^{D}$ are trained on support density features using the EM algorithm. Then these density multi prototypes‌ and ‌query features‌ are employed to encode the ‌local density similarity matrix‌ in the multiple local density learner. These local density similarity matrixes‌ are respectively concatenated with query features (U), thereby activating and facilitating ‌local guidance; In the global guidance‌, global density feature is encoded by summing the support density features. Then the global density and local-activation query features are used to guide the model in a global way, culminating in density map prediction for query image. In the inference phase, support density features‌ are encoded using ‌the above identical method, and the multiple local density learner is also optimized by the EM algorithm‌ to encode these local density similarity matrixes, subsequently enabling the ‌local-to-global density guidance‌ to predict the ‌density map‌ of query image from the unseen fixed surveillance scene.

\section{Experiments}
\subsection{Implementation}

To verify the efficacy of proposed framework, extensive experiments are performed across three benchmark surveillance datasets including WorldExpo'10 \cite{zhang2015cross}, Venice \cite{2019Context}, and CityUHK-X \cite{2020Incorporating}‌. Due to the ‌robust transfer learning capabilities‌, we employ a truncated VGG16 architecture retaining the initial 13 convolutional layers as the feature extraction backbone. During training, ‌data augmentation‌s are implemented through ‌mirroring operations‌ and ‌Gaussian blur kernels‌. Furthermore, patches with $257 \times 257 $ pixels are randomly sampled from original images. The adam optimization with ploy policy utilizes a ‌learning rate of 3e-6‌ and ‌batch size of 46‌. And the Euclidean distance loss is employed to optimize the predicted density map of query image. 

The ‌performance of the proposed method is ‌evaluated using the Mean Absolute Error (MAE) and Mean Squared Error (MSE)‌. They are defined as follows:

\begin{equation} MAE= \frac{1}{N}\sum _{n=1}^{N}\left | z_{n}-\widetilde{z_{n}}  \right |, \end{equation}

\begin{equation} MSE = \sqrt{\frac{1}{N}\sum ^{N}_{n=1}\left ( z_{n} - \widetilde{z_{n}} \right )^{2}}, \end{equation}where $N$ denotes the total number of images. $z_{n}$ denotes the actual crowd counting result, and $\widetilde{z_{n}}$ denotes the predicted crowd counting result.


\begin{table*}
\setlength{\abovecaptionskip}{0.12cm}
\setlength{\belowcaptionskip}{-0.35cm}
\caption{Comparison (MAE) with recent state-of-the-art methods on the WorldExpo'10 dataset.}
\begin{center}
\begin{tabular}{|p{1cm}|c|c|c|c|c|c|c|}\hline
\multicolumn{1}{|c|}{Method}                                & Scene 1 & Scene 2 & Scene 3 & Scene 4 & Scene 5 & \textcolor{green}{Average} \\ \hline
\hline








\multicolumn{1}{|c|}{CVPR2021 \cite{Ranjan_2021_CVPR}}                & 4.5   & 13.4  & 10.0   & 10.8  & 4.9   &  \textcolor{green}{8.7}  \\ \hline

\multicolumn{1}{|c|}{WACV2020 \cite{Reddy_2020_WACV}}                & 3.2   & 11.2  & 8.1   & 9.4  & 3.8   &  \textcolor{green}{7.1}  \\ \hline


\multicolumn{1}{|c|}{WACV2023 \cite{You_2023_WACV}}                & 3.2   & 11.0  & 8.3   & 9.1  & 3.6   &  \textcolor{green}{7.0}  \\ \hline

\multicolumn{1}{|c|}{IJCAI2024 \cite{xu2024learning}}                & 2.7   & 10.8  & 8.1   & 8.2  & 2.8   &  \textcolor{green}{6.5}  \\ \hline

\multicolumn{1}{|c|}{TIP2024 \cite{chen2024one}}               & 2.3   & 11.0  & 7.9   & 7.6  & 2.3   &  \textcolor{green}{6.2}  \\ \hline

\multicolumn{1}{|c|}{CVPR2025 \cite{chen2025fedbip}}                & 2.2   & 10.5  & 8.1   & 7.7  & 2.6   &  \textcolor{green}{6.2}  \\ \hline

\multicolumn{1}{|c|}{CVPR2025 \cite{chen2025single}}                & 2.2   & 10.1  & 7.7   & 7.8  & 2.4   &  \textcolor{green}{6.0}  \\ \hline


\multicolumn{1}{|c|}{LGD-OSCC}          &\textcolor{red}{\textbf{2.1}}   &\textcolor{red}{\textbf{10.0}}  &  \textcolor{red}{\textbf{7.5}}  & \textcolor{red}{\textbf{7.4}}  & \textcolor{red}{\textbf{2.2}}   &\textcolor{red}{\textbf{5.8}}  \\ \hline
\end{tabular}
\end{center}
\label{Table:4}
\end{table*}




\begin{table}
\setlength{\abovecaptionskip}{0.2cm}
\setlength{\belowcaptionskip}{-0.01cm}
\centering
\caption{Comparison with recent state-of-the-art methods on the Venice dataset and UHK dataset.}

\resizebox{.48\textwidth}{!}{ 
\begin{tabular}{|c|c|c|c|}
\hline
\multicolumn{1}{|c|}{\multirow{2}{*}{Methods}}        & \multicolumn{2}{c|}{Venice}       & \multicolumn{1}{c|}{UHK}  \\  \cline{2-4}    
                                  & MAE           & MSE              & MAE                        \\  \hline


CVPR2021 \cite{Ranjan_2021_CVPR}      & 22.3          & 32.4            & 7.4                             \\  \hline

WACV2020 \cite{Reddy_2020_WACV}          & 18.2          & 26.5            & 5.8                             \\  \hline

WACV2023 \cite{You_2023_WACV}          & 17.8          & 25.8            & 5.6                             \\  \hline

IJCAI2024 \cite{xu2024learning}          & 14.4          & 23.6            & 3.6                             \\  \hline


TIP2024 \cite{chen2024one}          & 13.9          & 20.0            & 2.7                             \\  \hline

CVPR2025 \cite{chen2025single}          & 13.0          & 18.6            & 2.4                             \\  \hline

CVPR2025 \cite{chen2025fedbip}          & 12.6         & 19.0            & 2.5                             \\  \hline

LGD-OSCC                   & \textbf{12.4}    & \textbf{18.0}          & \textbf{2.1}                 \\  \hline

\end{tabular}

}
\label{Table:3}
\end{table}

\begin{figure}
\setlength{\abovecaptionskip}{0.1cm}
\setlength{\belowcaptionskip}{-0.3cm}
\centering
   \includegraphics[width=1.0\linewidth]{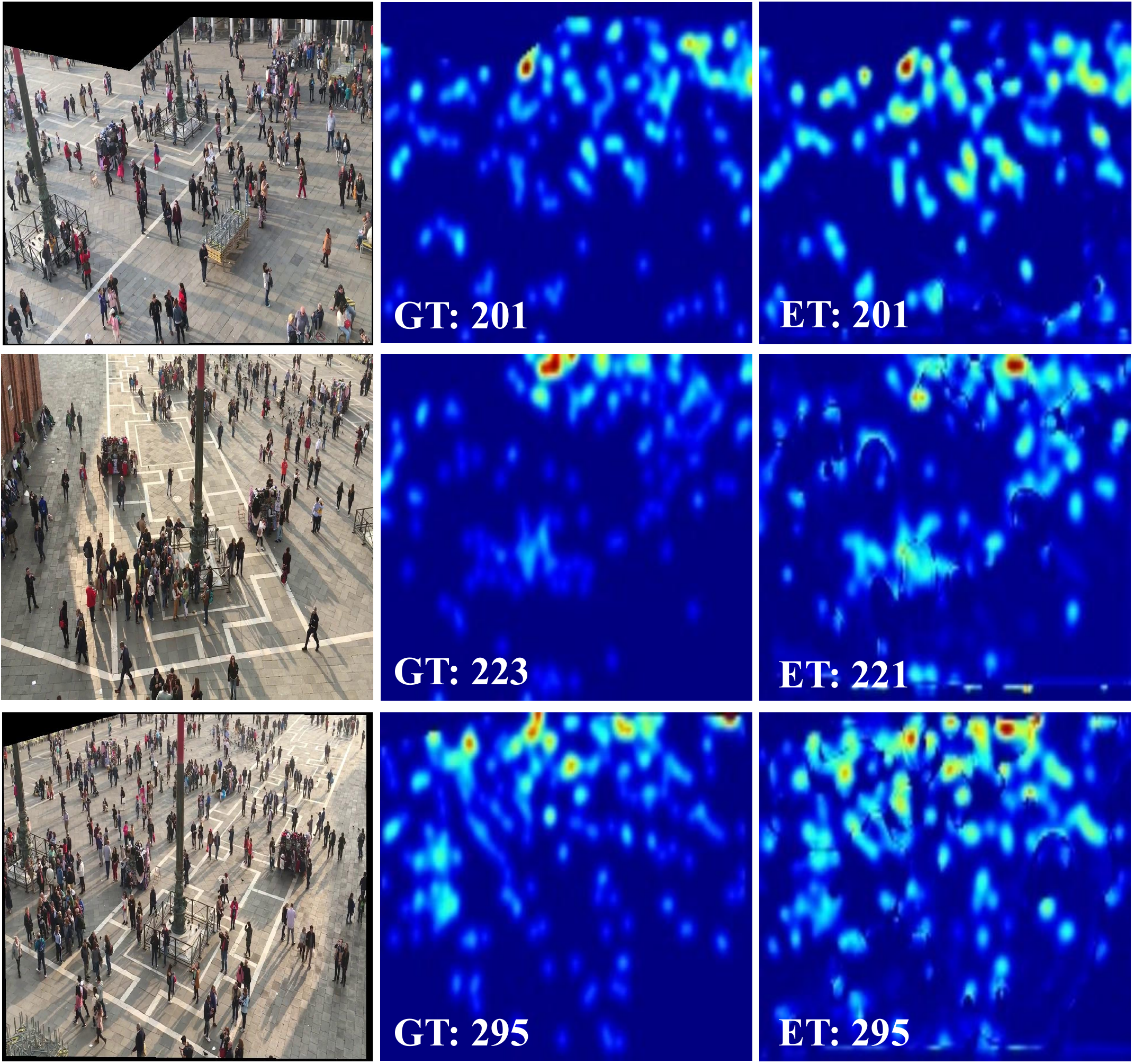}

   \caption{Visualization results on the Venice. First column: test images from three unseen surveillance scenes. Second column: ground-truth density map. Third column: the predicted query density map by the proposed LGD-OSCC. GT represents the ground-truth counting result. ET represents the estimated counting result.}
\label{FIG:5}
\end{figure}

\subsection{Comparisons with State-of-the-art Methods}

\noindent{\bfseries WorldExpo'10 \cite{zhang2015cross}.} The WorldExpo'10 benchmark dataset comprises 3,980 annotated images (resolution: $576 \times 720$ pixels), systematically subsampled from 1,132 video sequences captured across 108 different surveillance scenes. Each scene is accompanied by a predefined region of interest. Both original images and their corresponding ground-truth density maps are masked with the region of interest to ensure spatial consistency. The training set contains 3,380 images covering 103 surveillance scenes. The test set contains 600 images covering 5 surveillance scenes (designated as Scene 1 through Scene 5). Following the work \cite{chen2024one}, ground-truth density maps are generated through bivariate normal distribution modeling, where both pedestrian heads and bodies are annotated using the normalized Gaussian kernel. This standardized encoding facilitates fair cross-method comparisons in crowd counting research.

Several recent state-of-the-art few-shot learning methods have been comprehensively reimplemented in Table \ref{Table:4}. Experimental results demonstrate that our proposed method consistently achieves superior performance compared to the second best method \cite{chen2025single} across five evaluation scenes. Specifically, the proposed LGD-OSCC demonstrates performance improvements of $5\%$, $1\%$, $3\%$, $5\%$, and $8\%$ across five different evaluation scenes over \cite{chen2025single}, which demonstrates strong generalization ability across different crowd scenes. Hence, the proposed LGD-OSCC achieves the superior performance in terms of the average of five different surveillance scenes.


\begin{figure}[h!]
\centering
   \includegraphics[width=1.0\linewidth]{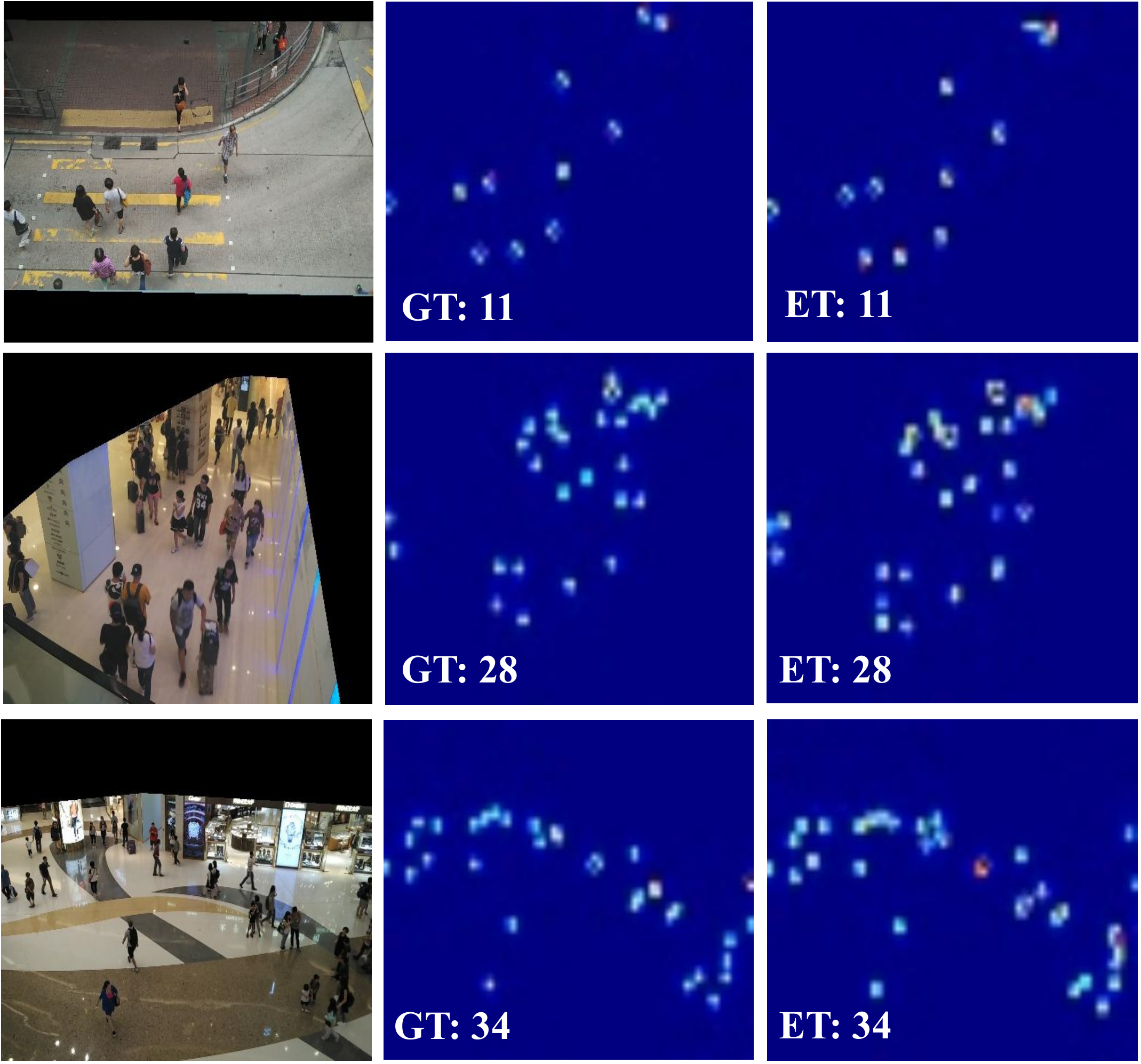}

   \caption{Visualization results on the CityUHK-X. First column: test images from three unseen surveillance scenes. Second column: ground-truth density map. Third column: the predicted query density map by the proposed LGD-OSCC. GT represents the ground-truth counting result. ET represents the estimated counting result.}
\label{FIG:6}
\end{figure}

\noindent{\bfseries Venice \cite{2019Context}.} The Venice dataset presents a novel challenge for surveillance-based crowd counting, featuring a single surveillance scene training set (80 images) and a multi-scene test set (3 different surveillance scenes, 87 images). All 167 images maintain a consistent resolution of $1,280 \times 720$ pixels, with the test set encompassing three different surveillance scenes to evaluate the generalization of methods.

A systematic comparison against these state-of-the-art methods is presented in Table \ref{Table:3}, where the performance through MAE/MSE averages across three different test scenes is reported. We can observe that the proposed LGD-OSCC achieves a $2\%$ MAE reduction versus the second best method \cite{chen2025fedbip}, despite being trained on singular-scene data. This consistent cross-domain superiority further confirms the exceptional generalization capabilities of our proposed method. These predicted density maps of multi-scene test set are shown in Figure \ref{FIG:5}. It can be found that these results further confirm the method's generalization capability across different surveillance scenes.



\noindent{\bfseries CityUHK-X \cite{2020Incorporating}.} The CityUHK-X (UHK) dataset comprises 3,191 surveillance images captured across 55 different surveillance scenes, and their resolution is $512 \times 384$. Camera configurations span inclination angles of $[-10^{\circ},-65^{\circ}]$ and installation heights ranging from $[2.2,16.0]$ meters. This dataset is systematically partitioned into two distinct subsets: a ‌training set‌ comprising ‌2,503 images‌ collected from ‌43 different surveillance scenes‌, and a ‌test set‌ consisting of ‌688 images‌ collected from ‌12 different surveillance scenes‌. This partitioning strategy is  designed to ensure that the model's generalizability is tested across a variety of unseen environments. 




Comparative performance against state-of-the-art methods is quantified in Table \ref{Table:3}. Following the work \cite{2020Incorporating}, MAE is used as the primary evaluation metric. Despite significant heterogeneity in camera installation angles and heights of this dataset, the proposed LGD-OSCC demonstrates superior generalizability, achieving a $13\%$  MAE reduction over the nearest competitor \cite{chen2025single}. This performance advantage is further corroborated by normalized visualizations across different surveillance scenes (Figure \ref{FIG:6}), which exhibit near-optimal crowd density estimation fidelity.

\begin{table}
\setlength{\abovecaptionskip}{0.2cm}
\centering
\caption{Estimation errors of the number ($\dot{v}$) of prototypes.}

  \begin{tabular}{|c|c|c|c|c|}
\hline
\multicolumn{1}{|c|}{\multirow{2}{*}{K}}        & \multicolumn{2}{c|}{WorldExpo'10}   & \multicolumn{2}{c|}{Venice}  \\  \cline{2-5}    
                                  & MAE           & MSE              & MAE           & MSE              \\  \hline
1                     & 7.4         & 9.7            & 19.1          & 25.5                    \\  \hline
2                     & 6.2      & 8.3            & 14.1          & 22.4                    \\  \hline
3                     & \textbf{5.8} &\textbf{8.0} &\textbf{12.4}   &\textbf{18.0}      \\  \hline
4                     & 6.1          & 8.4       & 12.8          & 19.6                    \\  \hline
5                     & 6.3    & 8.2              & 13.4         & 20.4                  \\  \hline

\end{tabular}
\label{Table:1}

\end{table}

\subsection{Ablation Study}

To systematically evaluate the contribution of individual architectural components including density multi prototypes, local density guidance, and global density guidance, comprehensive ablation experiments with analysis are conducted on two challenging benchmark datasets: WorldExpo'10 \cite{zhang2015cross} and Venice \cite{2019Context}.


\noindent{\bfseries Contribution of the density multi prototypes.} To systematically validate the advantages of density multi prototypes, comparative experiments are conducted across both benchmark datasets with varying numbers of prototypes $\dot{v}$. As demonstrated in Table \ref{Table:1}, there are statistically significant performance disparity between single prototype ($\dot{v}=1$) and multi prototypes ($\dot{v}>1$), justifying the necessity of multi-prototype integration. The best performance is achieved at $\dot{v}=3$, establishing this as the default configuration. Notably, performance degradation observed at higher values ($\dot{v}=4,5$) can be attributed to overfitting risks induced by excessive prototype diversification under the constraint of limited one-shot support image. This phenomenon underscores the importance of balanced selection in the number of density multi prototypes for robust feature representation.


\begin{table}
\setlength{\abovecaptionskip}{0.2cm}
\centering
\caption{Estimation errors of the Dilation Rate (\textbf{DR}) in Conv3.}

        \begin{tabular}{|c|c|c|c|c|}
\hline
\multicolumn{1}{|c|}{\multirow{2}{*}{DR}}        & \multicolumn{2}{c|}{WorldExpo'10}       & \multicolumn{2}{c|}{Venice}  \\  \cline{2-5}    
                                  & MAE           & MSE              & MAE           & MSE              \\  \hline
1                 & 6.1         & 8.5            & 12.8          & 19.5                    \\  \hline
2                      & \textbf{5.8}    & \textbf{8.0}          & \textbf{12.4}    & \textbf{18.0}            \\  \hline
3                   & 6.3    & 8.5          & 13.0    & 19.6                     \\  \hline

\end{tabular}
\label{Table:01}

\end{table}

\begin{table}
\setlength{\abovecaptionskip}{0.2cm}
\centering
\caption{Estimation errors of the local-to-global density guidance.}

\begin{tabular}{|c|c|c|c|c|}
\hline
\multicolumn{1}{|c|}{\multirow{2}{*}{Methods}}        & \multicolumn{2}{c|}{WorldExpo'10}       & \multicolumn{2}{c|}{Venice}  \\  \cline{2-5}    
                                  & MAE           & MSE              & MAE           & MSE              \\  \hline
W/o LDG                     & 7.7         & 11.8            & 17.6          & 24.3                    \\  \hline
W/o GDG                     & 6.5          & 10.0            & 15.3          & 20.2                    \\  \hline
LGD-OSCC                    & \textbf{5.8}    & \textbf{8.0}          & \textbf{12.4}    & \textbf{18.0}                    \\  \hline

\end{tabular}
\label{Table:2}

\end{table}

\noindent{\bfseries Contribution of the local density guidance.} To demonstrate the efficacy of local density guidance, we conduct two comparative experiments with analysis. The first experimental series examine the impact of varying dilation rates in ‌convolution-activation layer (Conv3), with quantitative results presented in Table \ref{Table:01}. The best performance is achieved at a dilation rate of 2, suggesting this particular receptive field configuration demonstrates superior efficacy in crowd feature extraction. The second experimental series involve the ablation of the local density guidance, and the Local Density Guidance (LDG) is removed from the LGD-OSCC. As shown in Table \ref{Table:2}, our proposed LGD-OSCC framework exhibits significant performance advantages (Table \ref{Table:2}, row 2 VS row 4), achieving MAE/MSE reductions of 1.9/3.8 on the WorldExpo'10 dataset and 5.2/6.3 on the Venice dataset. These results substantiate LDG's critical role in the generalization of models for crowd counting tasks.


\noindent{\bfseries Contribution of the global density guidance.} To validate the efficacy of global density guidance, comparative experiments are also conducted on the WorldExpo'10 dataset and Venice dataset. The Global Density Guidance (GDG) is removed from the proposed LGD-OSCC. Comparative results (Table \ref{Table:2}, row 3 VS row 4) demonstrate that removing the transformer-based GDG module (W/o GDG) significantly degrades performance compared to the complete LGD-OSCC framework. As anticipated, the GDG critically enhances the generalization of models, exhibiting MAE/MSE reductions of 0.7/2.0 on WorldExpo'10 and 2.9/2.2 on Venice, confirming its vital role in mitigating crowd counting errors.


\section{Conclusion}

In this work, we propose a novel density-guided one-shot crowd counting method to make model adapt to unseen surveillance scenes. To extract both local and global density features from the target scene, the ground-truth density map of support image is mapped to the support feature. Based on the mapping results, the global density feature can be obtained by summation, while local density similarity matrixes are encoded in the proposed multiple local density learner. To make the model adapt to the target surveillance scene, we propose a ‌local-to-global density-guided algorithm‌. Specifically, these local density similarity matrixes are first concatenated with the query feature and activated via the CNN, thereby performing ‌local density guidance‌. Subsequently, ‌the global density guidance‌ is implemented by leveraging the transformer and global density feature. Comprehensive experiments are performed on three benchmark surveillance datasets. These results conclusively indicate the superior generalization and performance over these state-of-the-art methods in few-shot crowd counting.

\noindent{\bfseries Acknowledgments} 
\newline This work is supported by the Natural Science Research Project of Higher Education Institutions in Jiangsu (No.25KJB510013), Jing-Jin-Ji Regional Integrated Environmental Improvement-National Science and Technology (No.2025ZD1201900), and the Startup Foundation for Introducing Talent of NUIST. We also acknowledge the High Performance Computing Center of Nanjing University of Information Science \& Technology for providing computational support.

\bibliographystyle{named}
\bibliography{mulu_13_1}

@inproceedings{zhang2016single,
  title={Single-image crowd counting via multi-column convolutional neural network},
  author={Zhang, Yingying and Zhou, Desen and Chen, Siqin and Gao, Shenghua and Ma, Yi},
  booktitle={Proceedings of the IEEE conference on computer vision and pattern recognition},
  pages={589--597},
  year={2016}
}

@inproceedings{benenson2014ten,
  title={Ten years of pedestrian detection, what have we learned?},
  author={Benenson, Rodrigo and Omran, Mohamed and Hosang, Jan and Schiele, Bernt},
  booktitle={European Conference on Computer Vision},
  pages={613--627},
  year={2014},
  organization={Springer}
}

@inproceedings{babu2018divide,
  title={Divide and Grow: Capturing Huge Diversity in Crowd Images With Incrementally Growing CNN},
  author={Babu Sam, Deepak and Sajjan, Neeraj N and Venkatesh Babu, R and Srinivasan, Mukundhan},
  booktitle={Proceedings of the IEEE Conference on Computer Vision and Pattern Recognition},
  pages={3618--3626},
  year={2018}
}

@inproceedings{zhang2015cross,
  title={Cross-scene crowd counting via deep convolutional neural networks},
  author={Zhang, Cong and Li, Hongsheng and Wang, Xiaogang and Yang, Xiaokang},
  booktitle={Proceedings of the IEEE conference on computer vision and pattern recognition},
  pages={833--841},
  year={2015}
}

@inproceedings{li2008estimating,
  title={Estimating the number of people in crowded scenes by mid based foreground segmentation and head-shoulder detection},
  author={Li, Min and Zhang, Zhaoxiang and Huang, Kaiqi and Tan, Tieniu},
  booktitle={2008 19th International Conference on Pattern Recognition},
  pages={1--4},
  year={2008},
  organization={IEEE}
}

@inproceedings{zeng2010robust,
  title={Robust head-shoulder detection by pca-based multilevel hog-lbp detector for people counting},
  author={Zeng, Chengbin and Ma, Huadong},
  booktitle={2010 20th International Conference on Pattern Recognition},
  pages={2069--2072},
  year={2010},
  organization={IEEE}
}

@inproceedings{Liu2018Crowd,
  title={Crowd Counting using Deep Recurrent Spatial-Aware Network},
  author={Liu, Lingbo and Wang, Hongjun and Li, Guanbin and Ouyang, Wanli and Liang, Lin},
  booktitle={Twenty-Seventh International Joint Conference on Artificial Intelligence {IJCAI-18}},
  year={2018},
}

@article{CHEN2019,
  title={Crowd counting with crowd attention convolutional neural network},
  author={ Chen, Jiwei  and  Wen, Su  and  Wang, Zengfu },
  journal={Neurocomputing},
  volume={382},
  pages={210--220},
  year={2020},
  publisher={Elsevier}

}

@inproceedings{LiuZPLDL18,
  author    = {Lingbo Liu and
               Ruimao Zhang and
               Jiefeng Peng and
               Guanbin Li and
               Bowen Du and
               Liang Lin},
  title     = {Attentive Crowd Flow Machines},
  booktitle = {2018 {ACM} Multimedia Conference on Multimedia Conference, {MM} 2018,
               Seoul, Republic of Korea, October 22-26, 2018},
  pages     = {1553--1561},
  publisher = {{ACM}},
  year      = {2018}
}

@article{gao2020cnn,
  title={Cnn-based density estimation and crowd counting: A survey},
  author={Gao, Guangshuai and Gao, Junyu and Liu, Qingjie and Wang, Qi and Wang, Yunhong},
  journal={arXiv preprint arXiv:2003.12783},
  year={2020}
}

@InProceedings{Reddy_2020_WACV,
author = {Reddy, Mahesh Kumar Krishna and Hossain, Mohammad and Rochan, Mrigank and Wang, Yang},
title = {Few-Shot Scene Adaptive Crowd Counting Using Meta-Learning},
booktitle = {Proceedings of the IEEE/CVF Winter Conference on Applications of Computer Vision (WACV)},
month = {March},
year = {2020}
}

@inproceedings{zhang2019canet,
  title={Canet: Class-agnostic segmentation networks with iterative refinement and attentive few-shot learning},
  author={Zhang, Chi and Lin, Guosheng and Liu, Fayao and Yao, Rui and Shen, Chunhua},
  booktitle={Proceedings of the IEEE/CVF Conference on Computer Vision and Pattern Recognition},
  pages={5217--5226},
  year={2019}
}

@inproceedings{yang2020prototype,
  title={Prototype mixture models for few-shot semantic segmentation},
  author={Yang, Boyu and Liu, Chang and Li, Bohao and Jiao, Jianbin and Ye, Qixiang},
  booktitle={European Conference on Computer Vision},
  pages={763--778},
  year={2020},
  organization={Springer}
}

@article{2020Incorporating,
  title={Incorporating Side Information by Adaptive Convolution},
  author={ Kang, D.  and  Chan, A. B. },
  journal={International Journal of Computer Vision},
  year={2020},
}

@inproceedings{2019Context,
  title={Context-Aware Crowd Counting},
  author={ Liu, W.  and  Salzmann, M.  and  Fua P. },
  booktitle={2019 IEEE/CVF Conference on Computer Vision and Pattern Recognition (CVPR)},
  year={2019},
}

@inproceedings{2018Few,
  title={Few-Shot Semantic Segmentation with Prototype Learning},
  author={ Dong, Nanqing  and  Xing, Eric P },
  booktitle={British Machine Vision Conference},
  year={2018},
}

@inproceedings{Sachin2017,
  title={OPTIMIZATION AS A MODEL FOR FEW-SHOT LEARNING},
  author={Sachin Ravi and Hugo Larochelle},
  booktitle={International Conference on Learning Representations},
  year={2017},
}

@article{CHEN2023109506, title = {Multi-task semi-supervised crowd counting via global to local self-correction}, journal = {Pattern Recognition}, volume = {140}, pages = {109506}, year = {2023}, issn = {0031-3203}, doi = {https://doi.org/10.1016/j.patcog.2023.109506}, author = {Jiwei Chen and Zengfu Wang} }

@ARTICLE{9161353,   author={Mo, Hong and Ren, Wenqi and Xiong, Yuan and Pan, Xiaoqi and Zhou, Zhong and Cao, Xiaochun and Wu, Wei},   journal={IEEE Transactions on Image Processing},    title={Background Noise Filtering and Distribution Dividing for Crowd Counting},    year={2020},   volume={29},   number={},   pages={8199-8212},   doi={10.1109/TIP.2020.3009030}}

@InProceedings{You_2023_WACV,
    author    = {You, Zhiyuan and Yang, Kai and Luo, Wenhan and Lu, Xin and Cui, Lei and Le, Xinyi},
    title     = {Few-Shot Object Counting With Similarity-Aware Feature Enhancement},
    booktitle = {Proceedings of the IEEE/CVF Winter Conference on Applications of Computer Vision (WACV)},
    month     = {January},
    year      = {2023},
    pages     = {6315-6324}
}

@InProceedings{Ranjan_2021_CVPR,
    author    = {Ranjan, Viresh and Sharma, Udbhav and Nguyen, Thu and Hoai, Minh},
    title     = {Learning To Count Everything},
    booktitle = {Proceedings of the IEEE/CVF Conference on Computer Vision and Pattern Recognition (CVPR)},
    month     = {June},
    year      = {2021},
    pages     = {3394-3403}
}

@InProceedings{Liu_2022_CVPR,
    author    = {Liu, Jie and Bao, Yanqi and Xie, Guo-Sen and Xiong, Huan and Sonke, Jan-Jakob and Gavves, Efstratios},
    title     = {Dynamic Prototype Convolution Network for Few-Shot Semantic Segmentation},
    booktitle = {Proceedings of the IEEE/CVF Conference on Computer Vision and Pattern Recognition (CVPR)},
    month     = {June},
    year      = {2022},
    pages     = {11553-11562}
}

@inproceedings{liang2023crowdclip,
  title={Crowdclip: Unsupervised crowd counting via vision-language model},
  author={Liang, Dingkang and Xie, Jiahao and Zou, Zhikang and Ye, Xiaoqing and Xu, Wei and Bai, Xiang},
  booktitle={Proceedings of the IEEE/CVF conference on computer vision and pattern recognition},
  pages={2893--2903},
  year={2023}
}

@InProceedings{Paiss_2023_ICCV,
    author    = {Paiss, Roni and Ephrat, Ariel and Tov, Omer and Zada, Shiran and Mosseri, Inbar and Irani, Michal and Dekel, Tali},
    title     = {Teaching CLIP to Count to Ten},
    booktitle = {Proceedings of the IEEE/CVF International Conference on Computer Vision (ICCV)},
    month     = {October},
    year      = {2023},
    pages     = {3170-3180}
}

@inproceedings{kang2024vlcounter,
  title={Vlcounter: Text-aware visual representation for zero-shot object counting},
  author={Kang, Seunggu and Moon, WonJun and Kim, Euiyeon and Heo, Jae-Pil},
  booktitle={Proceedings of the AAAI Conference on Artificial Intelligence},
  volume={38},
  number={3},
  pages={2714--2722},
  year={2024}
}

@inproceedings{wang2024language,
  title={Language-Guided Zero-Shot Object Counting},
  author={Wang, Mingjie and Yuan, Song and Li, Zhuohang and Zhu, Longlong and Buys, Eric and Gong, Minglun},
  booktitle={2024 IEEE International Conference on Multimedia and Expo Workshops (ICMEW)},
  pages={1--6},
  year={2024},
  organization={IEEE}
}

@article{ren2024grounding,
  title={Grounding dino 1.5: Advance the" edge" of open-set object detection},
  author={Ren, Tianhe and Jiang, Qing and Liu, Shilong and Zeng, Zhaoyang and Liu, Wenlong and Gao, Han and Huang, Hongjie and Ma, Zhengyu and Jiang, Xiaoke and Chen, Yihao and others},
  journal={arXiv preprint arXiv:2405.10300},
  year={2024}
}

@inproceedings{yao2024detclipv3,
  title={Detclipv3: Towards versatile generative open-vocabulary object detection},
  author={Yao, Lewei and Pi, Renjie and Han, Jianhua and Liang, Xiaodan and Xu, Hang and Zhang, Wei and Li, Zhenguo and Xu, Dan},
  booktitle={Proceedings of the IEEE/CVF conference on computer vision and pattern recognition},
  pages={27391--27401},
  year={2024}
}

@inproceedings{chen2025single,
  title={Single Domain Generalization for Few-Shot Counting via Universal Representation Matching},
  author={Chen, Xianing and Huo, Si and Jiang, Borui and Hu, Hailin and Chen, Xinghao},
  booktitle={Proceedings of the Computer Vision and Pattern Recognition Conference},
  pages={4639--4649},
  year={2025}
}

@inproceedings{li2018learning,
  title={Learning to generalize: Meta-learning for domain generalization},
  author={Li, Da and Yang, Yongxin and Song, Yi-Zhe and Hospedales, Timothy},
  booktitle={Proceedings of the AAAI conference on artificial intelligence},
  volume={32},
  number={1},
  year={2018}
}

@inproceedings{zhu2023daot,
  title={DAOT: Domain-agnostically aligned optimal transport for domain-adaptive crowd counting},
  author={Zhu, Huilin and Yuan, Jingling and Zhong, Xian and Yang, Zhengwei and Wang, Zheng and He, Shengfeng},
  booktitle={Proceedings of the 31st ACM International Conference on Multimedia},
  pages={4319--4329},
  year={2023}
}

@inproceedings{du2023domain,
  title={Domain-general crowd counting in unseen scenarios},
  author={Du, Zhipeng and Deng, Jiankang and Shi, Miaojing},
  booktitle={Proceedings of the AAAI conference on artificial intelligence},
  volume={37},
  number={1},
  pages={561--570},
  year={2023}
}

@InProceedings{Peng_2024_CVPR,
    author    = {Peng, Zhuoxuan and Chan, S.-H. Gary},
    title     = {Single Domain Generalization for Crowd Counting},
    booktitle = {Proceedings of the IEEE/CVF Conference on Computer Vision and Pattern Recognition (CVPR)},
    month     = {June},
    year      = {2024},
    pages     = {28025-28034}
}

@inproceedings{choi2021robustnet,
  title={Robustnet: Improving domain generalization in urban-scene segmentation via instance selective whitening},
  author={Choi, Sungha and Jung, Sanghun and Yun, Huiwon and Kim, Joanne T and Kim, Seungryong and Choo, Jaegul},
  booktitle={Proceedings of the IEEE/CVF conference on computer vision and pattern recognition},
  pages={11580--11590},
  year={2021}
}

@inproceedings{pan2019switchable,
  title={Switchable whitening for deep representation learning},
  author={Pan, Xingang and Zhan, Xiaohang and Shi, Jianping and Tang, Xiaoou and Luo, Ping},
  booktitle={Proceedings of the IEEE/CVF international conference on computer vision},
  pages={1863--1871},
  year={2019}
}

@InProceedings{Qian_2025_CVPR,
    author    = {Qian, Yifei and Guo, Zhongliang and Deng, Bowen and Lei, Chun Tong and Zhao, Shuai and Lau, Chun Pong and Hong, Xiaopeng and Pound, Michael P.},
    title     = {T2ICount: Enhancing Cross-modal Understanding for Zero-Shot Counting},
    booktitle = {Proceedings of the IEEE/CVF Conference on Computer Vision and Pattern Recognition (CVPR)},
    month     = {June},
    year      = {2025},
    pages     = {25336-25345}
}

@InProceedings{Doubinsky_2024_WACV,
    author    = {Doubinsky, Perla and Audebert, Nicolas and Crucianu, Michel and Le Borgne, Herv\'e},
    title     = {Semantic Generative Augmentations for Few-Shot Counting},
    booktitle = {Proceedings of the IEEE/CVF Winter Conference on Applications of Computer Vision (WACV)},
    month     = {January},
    year      = {2024},
    pages     = {5443-5452}
}

@article{chen2024one,
  title={One-shot any-scene crowd counting with local-to-global guidance},
  author={Chen, Jiwei and Wang, Zengfu},
  journal={IEEE Transactions on Image Processing},
  volume={33},
  pages={6622--6632},
  year={2024},
  publisher={IEEE}
}

@inproceedings{wen2021detection,
  title={Detection, tracking, and counting meets drones in crowds: A benchmark},
  author={Wen, Longyin and Du, Dawei and Zhu, Pengfei and Hu, Qinghua and Wang, Qilong and Bo, Liefeng and Lyu, Siwei},
  booktitle={Proceedings of the IEEE/CVF Conference on Computer Vision and Pattern Recognition},
  pages={7812--7821},
  year={2021}
}

@inproceedings{chen2025fedbip,
  title={Fedbip: Heterogeneous one-shot federated learning with personalized latent diffusion models},
  author={Chen, Haokun and Li, Hang and Zhang, Yao and Bi, Jinhe and Zhang, Gengyuan and Zhang, Yueqi and Torr, Philip and Gu, Jindong and Krompass, Denis and Tresp, Volker},
  booktitle={Proceedings of the Computer Vision and Pattern Recognition Conference},
  pages={30440--30450},
  year={2025}
}

@inproceedings{xu2024learning,
  title={Learning spatial similarity distribution for few-shot object counting},
  author={Xu, Yuanwu and Song, Feifan and Zhang, Haofeng},
  booktitle={Proceedings of the Thirty-Third International Joint Conference on Artificial Intelligence},
  pages={1507--1515},
  year={2024}
}

@InProceedings{Pelhan_2024_CVPR,
    author    = {Pelhan, Jer and Luke\v{z}i, Alan and Zavrtanik, Vitjan and Kristan, Matej},
    title     = {DAVE - A Detect-and-Verify Paradigm for Low-Shot Counting},
    booktitle = {Proceedings of the IEEE/CVF Conference on Computer Vision and Pattern Recognition (CVPR)},
    month     = {June},
    year      = {2024},
    pages     = {23293-23302}
}

@article{gao2025survey,
  title={A survey of deep learning methods for density estimation and crowd counting},
  author={Gao, Guangshuai and Gao, Junyu and Liu, Qingjie and Wang, Qi and Wang, Yunhong},
  journal={Vicinagearth},
  volume={2},
  number={1},
  pages={1--37},
  year={2025},
  publisher={Springer}
}

@article{song2023comprehensive,
  title={A comprehensive survey of few-shot learning: Evolution, applications, challenges, and opportunities},
  author={Song, Yisheng and Wang, Ting and Cai, Puyu and Mondal, Subrota K and Sahoo, Jyoti Prakash},
  journal={ACM Computing Surveys},
  volume={55},
  number={13s},
  pages={1--40},
  year={2023},
  publisher={ACM New York, NY}
}


\end{document}